\documentclass[a4paper]{svproc}

\usepackage{xcolor}
\usepackage{color}

\usepackage{makeidx}         %
\usepackage{graphicx}        %
\usepackage{multicol}        %
\usepackage[bottom]{footmisc}%

\usepackage{amsmath}
\usepackage{amsfonts}
\usepackage{multirow}
\usepackage{tabularx}
\usepackage{booktabs}
\usepackage{hhline}
\usepackage{amsthm}
\usepackage{mathtools}
\usepackage{wrapfig}
\usepackage{lipsum}  
\usepackage{url}
\usepackage{subfig}
\usepackage{graphicx}

\usepackage[font={small}]{caption}

\definecolor{WildStrawberry}{rgb}{1.0, 0.26, 0.64}
\definecolor{ForestGreen}{rgb}{0.13, 0.55, 0.13}
\definecolor{teal}{RGB}{0, 161, 115}
\definecolor{brightBlue}{RGB}{0, 114, 207}
\definecolor{wine}{RGB}{207, 0, 48}

\newcommand{\klnote}[1]{\textcolor{black} {#1}}

\newcommand{\uA}{u_\mathrm{A}}    %
\newcommand{\uB}{u_\mathrm{B}}    %

\newcommand{\UA}{\mathcal{U}^\mathrm{A}}    %
\newcommand{\UB}{\mathcal{U}^\mathrm{B}}    %

\newcommand\blfootnote[1]{%
  \begingroup
  \renewcommand\thefootnote{}\footnote{#1}%
  \addtocounter{footnote}{-1}%
  \endgroup
}

\renewcommand{\baselinestretch}{0.96}

\begin{document}
\title{Towards Data-Driven Synthesis of\\Autonomous Vehicle Safety Concepts}
\titlerunning{Towards Data-Driven Synthesis of Autonomous Vehicle Safety Concepts}

\author{
  Karen Leung$^{1\star}$, Andrea Bajcsy$^{2\star}$, Edward Schmerling$^1$, Marco Pavone$^1$\\
}
\authorrunning{Karen Leung$^{\star}$, Andrea Bajcsy$^{\star}$, Edward Schmerling, Marco Pavone}
\institute{
NVIDIA, \texttt{\{kaleung,eschmerling,mpavone\}@nvidia.com}
\and University of California, Berkeley, \texttt{abajcsy@berkeley.edu}}
\maketitle

\begin{abstract}
As safety-critical autonomous vehicles (AVs) will soon become pervasive in our society, a number of \textit{safety concepts} for trusted AV deployment have recently been proposed throughout industry and academia. Yet, achieving consensus on an appropriate safety concept is still an elusive task. 
In this paper, we advocate for the use of Hamilton-Jacobi (HJ) reachability as a unifying mathematical framework for comparing existing safety concepts, and through elements of this framework propose ways to tailor safety concepts (and thus expand their applicability) to scenarios with implicit expectations on agent behavior in a data-driven fashion.
Specifically, we show that
(i) existing predominant safety concepts can be embedded in the HJ reachability framework, thereby enabling a common language for comparing and contrasting modeling assumptions, and (ii) HJ reachability can serve as an inductive bias to effectively reason, in a learning context, about two critical, yet often overlooked aspects of safety: responsibility and context-dependency. \blfootnote{$^\star$Indicates equal contribution. AB: Work performing during internship at NVIDIA.}
\keywords{autonomous vehicles, safe interaction, HJ reachability} 
\end{abstract}

\setcounter{footnote}{0}
\section{Introduction}
\klnote{Making a strong case for safety is an essential prerequisite for autonomous vehicle (AV) deployment and widespread adoption.
This is an objective at top of mind for AV system developers, policymakers, and the general public alike; while most would agree that \emph{guaranteeing} safety is not strictly possible in the face of the myriad uncertainties and complexities that come with real-world driving, there is still a broad desire to codify, at least for some situations, collectively agreed upon notions of safety \cite{ieee2022P2846,nister2019safety,shalev2017formal}.
Specifically towards understanding the aspects of safety pertaining to interactions with other road users, stakeholders including AV industry players, academics, and standards organizations have advanced various \emph{safety concepts} that aim to evaluate driving scenarios and decide what constitutes safe behavior.}
\klnote{The potential benefits are clear---over a set of simple interactive situations (currently, pairwise interactions with simplified road geometries and agent intent, but intended to be expanded over time), regulators have properties of AV behavior that they can demand and check, while engineers can ensure that these properties are embedded into the design of their AV systems.}
Underpinning these notions of safety are modeling assumptions reasoned from first principles or informed by data.
However, to date there is no universal agreement on how to select these assumptions and therefore on how and in which situations a consensus notion of safety can be assessed.\footnote{Unlike in the maritime or aviation domains where, e.g., the COLREGs \cite{colregs1972} and ACAS X \cite{kochenderfer2012next} specify universally agreed upon safety rules to be followed in all scenarios.}
In this paper, we do not purport to resolve which proposed safety concept, if any, is best. Instead, we aim to accelerate convergence to a shared consensus by casting these concepts within a common framework to promote comparison, novel synthesis, \klnote{and data-driven extension to more complex and nuanced driving scenarios}.

In this work, we define safety concepts as the combination of two functions mapping world state (e.g., joint state of all agents and environmental context like road geometry) to (i) a scalar measure of safety, and (ii) a set of allowable actions for each agent. We note that this definition abstracts away the computation by which modeling assumptions are transformed into such functions; indeed, factors such as measurement uncertainty, delayed reaction times, and possibly stochastic models of behavior are considered implicitly \klnote{through the two functions}. Aside from clear-cut cases where, for example, inevitable collision may be determined purely from dynamics considerations, deciding on a safety concept is challenging because of the \klnote{uncertainty in human behaviors, and more notably, the} necessary dependence on \emph{responsibility} (degree of each agent's ownership over collision-avoidance) as well as \emph{context}. In particular, safety and responsibility are inextricably entwined, whereby the safety of all but the most conservative maneuvers relies on the assumption that surrounding agents will act responsibly, while even defining responsibility requires a means to quantify how much agents' actions contribute to safety. The nuance of how safety and responsibility are influenced by context (e.g., a road blockage may allow an agent to briefly cross into an oncoming traffic lane which would otherwise be wholly irresponsible) motivates the usage of driving data in learning safety concepts. How to do this in a principled fashion, however, remains an open research question.

We propose that Hamilton-Jacobi (HJ) reachability is a promising unifying mathematical framework for describing safety concepts and provides, by design, the inductive bias that the scalar safety measure and allowable action sets should be consistent. 
In this work, we take the first steps to demonstrate that many seemingly disparate existing safety concepts can be unified via HJ reachability. We note that HJ reachability on its own does not answer the important question of which safety concept is most appropriate for AVs. Instead, it enables us to characterize a \textit{family of safety concepts} for which we may hope to design data-driven synthesis techniques for novel safety concepts. We further elaborate research directions on safety-centric dataset construction, interpretable representations of context, and learned notions of responsibility, highlighting the role of HJ reachabilty as inductive bias throughout.  
Our \klnote{ultimate} goal is to help stakeholders converge on a safety concept grounded in data, reflecting both realistic AV interactions while remaining interpretable from a policy perspective.

\section{A unifying framework for safe AV interactions}
\label{sec:2}
In this section, we provide a mathematical introduction to HJ reachability while highlighting key elements that make it possible to encompass a family of safety concepts and thus provide valuable inductive bias during data-driven synthesis and evaluation of safety concepts. 
By demonstrating the modeling flexibility of HJ reachability, we can more confidently take the first steps towards novel safety concepts which harness data to reason about responsibility and context-dependency.

\subsection{Background: Hamilton-Jacobi Reachability}
\label{subsec:hj background}
HJ reachability is a mathematical formalism for characterizing the performance and safety properties of (multi-agent) dynamical systems \cite{mitchell2005time,margellos2011hamilton,bansal2017hamilton}. Core to HJ reachability is the set of states $\mathcal{L}$ which agents reason about either seeking or avoiding within a time horizon $T$. In the context of collision avoidance between agents $A$ and $B$, $\mathcal{L}$ corresponds to the set of collision states. 
To capture the multi-agent, continuous-time, and safety-critical nature of our setting, HJ reachability describes a two-player
differential game. This formulation enables us to mathematically characterize whether it is possible for the ego agent to prevent an undesirable outcome under any family of
closed-loop policies of other agents, as well as the ego agent's appropriate control policy for ensuring safety.

Using the principle of dynamic programming, the collision avoidance problem between agents $A$ and $B$ reduces to solving the Hamilton-Jacobi-Isaacs (HJI) partial differential equation (PDE),
{\small
\begin{align}
    \frac{\partial V(z, t)}{\partial t} &+ 
    \min \Big\{0,~
    {\begingroup
    \overbrace{\textstyle {\color{brightBlue} \max}_{\uA \in {\color{WildStrawberry}\underbrace{\textstyle \UA}_{\text{A's control set}}}} {\color{brightBlue} \min}_{\uB \in {\color{WildStrawberry}\underbrace{\textstyle \UB}_{\text{B's control set}}}}}^{\color{brightBlue}\text{Type of agent's reactions}}
    \endgroup}\label{eq:hji_pde_brt}
    \nabla_{z} V(z, t)^\top f(z, \uA, \uB) \Big\} = 0 \\\notag
    V(z, 0) &= {\color{ForestGreen}\ell(z)\: \} \:\scriptstyle{\text{Safety criterion}}},
\end{align}
}
\normalsize
where $z \in \mathcal{Z}$ denotes the joint state of agents $A$ and $B$, $u_\mathrm{A}$ and $u_\mathrm{B}$ are the controls of agents $A$ and $B$, respectively, and $f(\cdot,\cdot,\cdot)$ is the joint dynamics. The boundary condition for this PDE is defined by the function ${\color{ForestGreen}\ell}: \mathcal{Z} \rightarrow \mathbb{R}$ whose zero sub-level set encodes the undesirable states $\mathcal{L}$, i.e., $\mathcal{L} =  \lbrace z \mid {\color{ForestGreen}\ell(z)} < 0\rbrace$.  Lastly, a more general reach-avoid formulation also exists, whereby the agents optimize with respect to both a goal set and avoid set. We refer the reader to \cite{margellos2011hamilton,fisac2014reachavoid} for more details.

By solving Equation~\eqref{eq:hji_pde_brt} \textit{backwards} in time over a time horizon of $T$, we obtain the HJ value function $V(z,t)$ for $t\in[-T,0]$. For any starting state $z \in \mathcal{Z}$, this function captures the closest the overall system can get to the set of undesirable states $\mathcal{L}$ (i.e., the lowest value of {\color{ForestGreen}$\ell(\cdot)$} along a system trajectory) within $|t|$ seconds if both agents $A$ and $B$ act optimally, $\uA^*(z),\,\uB^*(z) = \arg {\color{brightBlue}\max}_{\uA \in {\color{WildStrawberry}\UA}}{\color{brightBlue}\min}_{\uB \in {\color{WildStrawberry}\UB}} \nabla_{z} V(z, t)^\top f(z, \uA, \uB)$. %
Since \eqref{eq:hji_pde_brt} encodes a collision-avoidance problem, we can obtain 
the \textit{unsafe set} $\mathcal{A}(t)$ at time $t$ as the zero sub-level set of the value function:  $\mathcal{A}(t) = \lbrace z \mid V(z, t) < 0 \rbrace$.
\klnote{Moreoever, the \textit{safety-preserving control set}, $\mathcal{\tilde{U}}(\uB) = \lbrace \uA \in {\color{WildStrawberry}\UA} \mid \frac{\partial V(z, t)}{\partial t} + \nabla V(z,t)^\top f(z,\uA, \uB) \geq 0 \rbrace,$
is the set of controls that prevents the value (i.e., level of safety) from decreasing.}

HJ reachability thus synthesizes a safety measure through the value function $V$, which fulfills the first aspect of a safety concept. Importantly, this framework also facilitates \textit{learning} the second aspect---the allowable control sets---by ensuring consistency (in terms of dynamics and agent intent) between the control sets and their effect on the safety measure.
The family of safety concepts that can be encoded and learned in Equation~\eqref{eq:hji_pde_brt} ultimately depends on three key aspects of the HJ reachability formulation: (i) the type of agent reactions, (ii) allowable control sets, and (iii) the safety criterion.

\begin{figure}[t]
    \centering
    \includegraphics[width=\textwidth]{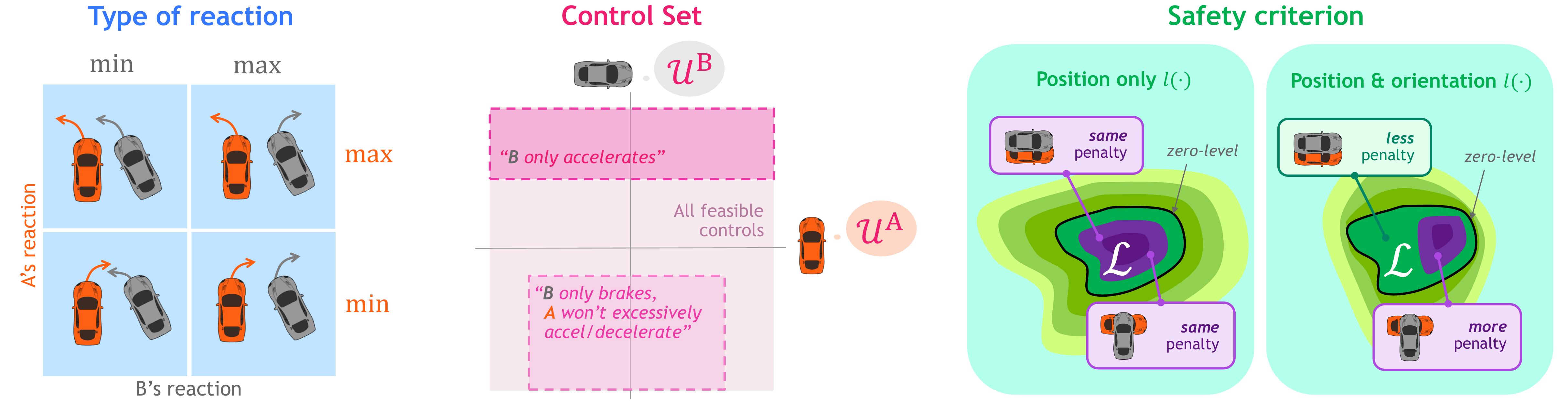}
    \caption{HJ reachability: allowing a spectrum of safety concepts in the same framework.}
    \label{fig:elements_of_hj}
\end{figure}

\noindent{\color{brightBlue} \textbf{Type of agents' reactions ($\max \text{ or } \min$)}}: 
As currently stated in \eqref{eq:hji_pde_brt}, agent $B$ has an informational advantage (by ``playing second'' and responding to $A$'s controls) and is modelled as acting \textit{adversarially} by \textit{minimizing} the instantaneous rate of change of the value function subject to the joint dynamics. 
By modifying the information pattern and/or if agents are minimizers or maximizers, we can modify the strategic optimism when computing the unsafe set (Figure~\ref{fig:elements_of_hj}, left).

\noindent{\color{WildStrawberry} \textbf{Allowable control sets ($\UA$ and $\UB$)}}: Traditionally, the control set represents feasible controls of each agent (e.g., actuation limits of the system). However, we can \textit{restrict} the control set (e.g., informed by data) to reflect assumptions about how other agents behave in safety-critical scenarios (Figure~\ref{fig:elements_of_hj}, center).

\noindent{\color{ForestGreen} \textbf{Safety criterion $\mathcal{L} = \lbrace z \mid \ell(z) < 0\rbrace$}}: We are free to define {\color{ForestGreen}$\ell(\cdot)$} as long as its zero sub-level set equals $\mathcal{L}$.
While purely geometric functions like penetration/separation distance are a common choice, we can design or learn alternative functions that capture more nuanced notions of safety. For example, by shaping {\color{ForestGreen}$\ell$} to penalize more dangerous orientations (e.g., T-bone or head-on collision) we can encode collision severity or collision responsibility (Figure~\ref{fig:elements_of_hj}, right). 

\subsection{Embedding existing safety concepts in HJ reachability}

Here, we briefly describe existing predominant safety concepts and demonstrate how they can be expressed via HJ reachability. We note that the majority of these concepts were developed without HJ reachability in mind. Throughout, we assume {\color{ForestGreen}$\ell(\cdot)$} encodes penetration/separation (i.e., signed) distance.

\noindent\textbf{Worst-case dynamic game \cite{mitchell2005time,leung2020infusing}}: An \textit{unsafe} state is one where a collision is inevitable despite agent $A$'s best effort to avoid collision and agent $B$'s policy to cause collision.
\textit{\textbf{HJ formulation}}: This worst-case {\color{brightBlue}$\max \min$} formulation is stated in \eqref{eq:hji_pde_brt}, allowing for {\color{WildStrawberry}all dynamically-feasible controls}. 

\noindent\textbf{Forward reachable set \cite{althoff2014online}}: The forward reachable set (FRS) is the set of states a system can reach at some future time, i.e., the FRS can be computed by simulating all possible control sequences a system can take to see which future states are possible. A safety check using FRS involves checking if the FRS of two agents intersect. \textit{\textbf{HJ formulation}}: Consider the {\color{brightBlue}$\min \min$} game where both agents optimize over {\color{WildStrawberry} any choice of dynamically-feasible controls}. Then $V(z,t) <0 \Leftrightarrow$ the intersection of the FRS of each agent is non-empty.

\noindent\textbf{Safety Force Field (SFF) \cite{nister2019safety}}: Assume each agent has a set of safety procedures, i.e., a set of control signals that brings an agent to a stop in finite time. An \textit{unsafe} state is one where there exists a pair of safety procedures which, if executed, lead to agents A and B colliding. \textit{\textbf{HJ formulation}}: This is a {\color{brightBlue}$\min \min$} game (since we want to detect a pair of safety procedures that lead to a collision) where both agents optimize over a restricted control set containing only the {\color{WildStrawberry}safety procedures}.

\noindent\textbf{Responsibility-Sensitive Safety (RSS) \cite{shalev2017formal}}:
An \textit{unsafe} state is one where a collision occurs if both agents apply a fixed deceleration in both the longitudinal and lateral dimension. 
\textit{\textbf{HJ formulation}}\footnote{For brevity, we ignore reaction time, but can account for this with a time-varying control set.}: Since longitudinal and lateral motions are decoupled, the corresponding unsafe set is the intersection of the longitudinal and lateral unsafe sets. 
To compute the longitudinal (lateral) unsafe set, we consider the longitudinal (lateral) dynamics and restrict the control set to only deceleration, i.e., {\color{WildStrawberry}$|\UA| = 1, |\UB| = 1$}.
With singleton control sets, the {\color{brightBlue}$\max \min$} operations can be removed.

\noindent\textbf{Contingency safety \cite{althoff2014online,kuwata2009real,chen2021guaranteed}}:
An \textit{unsafe} state is one where an agent is \textit{not} able to come to a collision-free stop assuming other agents maintain their current assumed motion.
The notion of ``contingency safety'' describes a class of safety concepts, depending on how other agents are assumed to behave. 
\textit{\textbf{HJ formulation}}: We may set up a reach-avoid game where agent $A$
strives to avoid agent $B$ while $A$ tries to reach a zero-velocity state. The type of game and agent $B$'s control set depends on the assumptions on its behavior (e.g., maintain speed in lane vs. hard brake). 

\noindent\textbf{Constant motion \cite{shiller1998motion,wilkie2009generalized}}:
An \textit{unsafe} state is one where collision occurs if both agents maintain their current velocity and steering. 
\textit{\textbf{HJ formulation}}: Augment the state-space with the control inputs and set control derivatives to zero. This state augmentation removes the game-theoretic aspect of the HJ formulation.

\section{Grounding responsibility and context-aware safety concepts in data}

The above safety concepts largely lack nuanced treatments of responsibility (instead rigidly assuming, for example, that agents follow prescribed policies in safety critical situations), and lack expressivity in incorporating local context. In this section we argue for the necessity of such considerations in deriving an ideal safety concept, and propose \klnote{a number of} research directions \klnote{\textbf{[D1]}--\textbf{[D5]}} according to the paradigm that context-dependency and the coupling between responsibility and safety should be grounded in data.

\subsection{Collecting datasets encompassing safety-critical situations}
What data is necessary to learn safety concepts? This question requires embarking on two research directions. First, \textbf{[D1]} developing new---or augmenting existing---datasets by actively querying humans for safety-relevant labels. We contend that logged trajectory data alone
may not capture counterfactual-dependent questions such as \textit{how safe is a scenario?}, \textit{did the agent(s) perform acceptable maneuvers?}, and, if a maneuver is labeled unsafe, \textit{what alternative(s) are preferable?} Second, \textbf{[D2]} datasets should ensure sufficient coverage of safety-critical scenarios while respecting a limited labeling budget. Here, policymakers can inform dataset coverage by enumerating a list of key AV scenario classes, while engineering stakeholders can reduce human effort through automated and targeted dataset augmentation; for example, by optimizing out-of-distribution metrics to capture epistemic uncertainty or taking inspiration from the Quality-Diversity optimization community \cite{chatzilygeroudis2021quality}. 

\subsection{Understanding how context influences safety}
Contextual reasoning has proven instrumental in advancing understanding in AV-relevant fields, e.g., as enabled in computer vision by the Common Objects in Context (COCO) dataset \cite{lin2014microsoft}. Towards capturing richer notions of AV safety without losing first-principles interpretability, we advocate for \textbf{[D3]} establishing representations of context that are tangible for stakeholders and operationalizable within safety computation. Incorporating context will be impactful in cases where even after accounting for road rules (which provide a relatively concrete assessment of safety when applicable) and controlling for differences in joint state, the data exhibits a high degree of variability in safety labels. As a concrete example, we may consider weather which influences visibility (e.g., foggy conditions) and maneuverability (e.g., icy roads) which in turn should impact both the safety measure and allowable action sets. In this case, we propose parameterizing the HJ reachability computation on the interpretable axes of visibility and maneuverability.

\subsection{Defining and inferring levels of responsibility of agents}
Consider a road blockage which motivates a brief excursion into oncoming traffic. 
Determining if this maneuver is allowable---let alone responsible---requires 
\textbf{[D4]} defining and inferring responsibility from context and possibly noisy data. 
Existing methods have leveraged noisy data to varying degrees in an effort to model and infer different notions of responsibility: as part of an agent's objective \cite{schwarting2019social,laine2020multi} or control space (i.e., how much ``effort'' an agent puts into collision-avoidance)  \cite{chen2021guaranteed,van2008reciprocal}. 
This naturally informs \textbf{[D5]} how to actionably incorporate responsibility into a safety concept. 
By ensuring that the safety measure and allowable control sets are consistent, HJ reachability serves as strong inductive bias when learning a cohesive combination of responsibility representations; for example, learning a restriction on the {\color{WildStrawberry}control spaces $\UA$ and $\UB$}, selecting appropriate {\color{brightBlue}agent reactions ($\max/\min$)}, or automatically ``shaping'' the {\color{ForestGreen}safety criterion $\ell(\cdot)$} via data. 

\section{Illustrative examples}
\klnote{
In this section we provide some examples to demonstrate how varying the elements of the reachability computation (see Section~\ref{subsec:hj background}) to reflect different behavioral assumptions (e.g., stemming from considerations of responsibility)  results in significantly different delineations between safe and unsafe states. In the following examples, we consider the relative state $z=[\Delta x, \Delta y, \Delta \theta, v_\mathrm{A}, v_\mathrm{B}]$ centered around Agent $A$, and relative dynamics derived from a four-state simple car model (controlled acceleration and steering angle). We use the signed distance function between car bodies to define ${\color{ForestGreen}\ell(z)}$ and consider a {\color{brightBlue}$\max \min$} formulation.}

\begin{figure}[t]
     \centering
     \subfloat[Comparing equal and unequal (responsibility-aware) control sets.]{\includegraphics[width=0.45\textwidth]{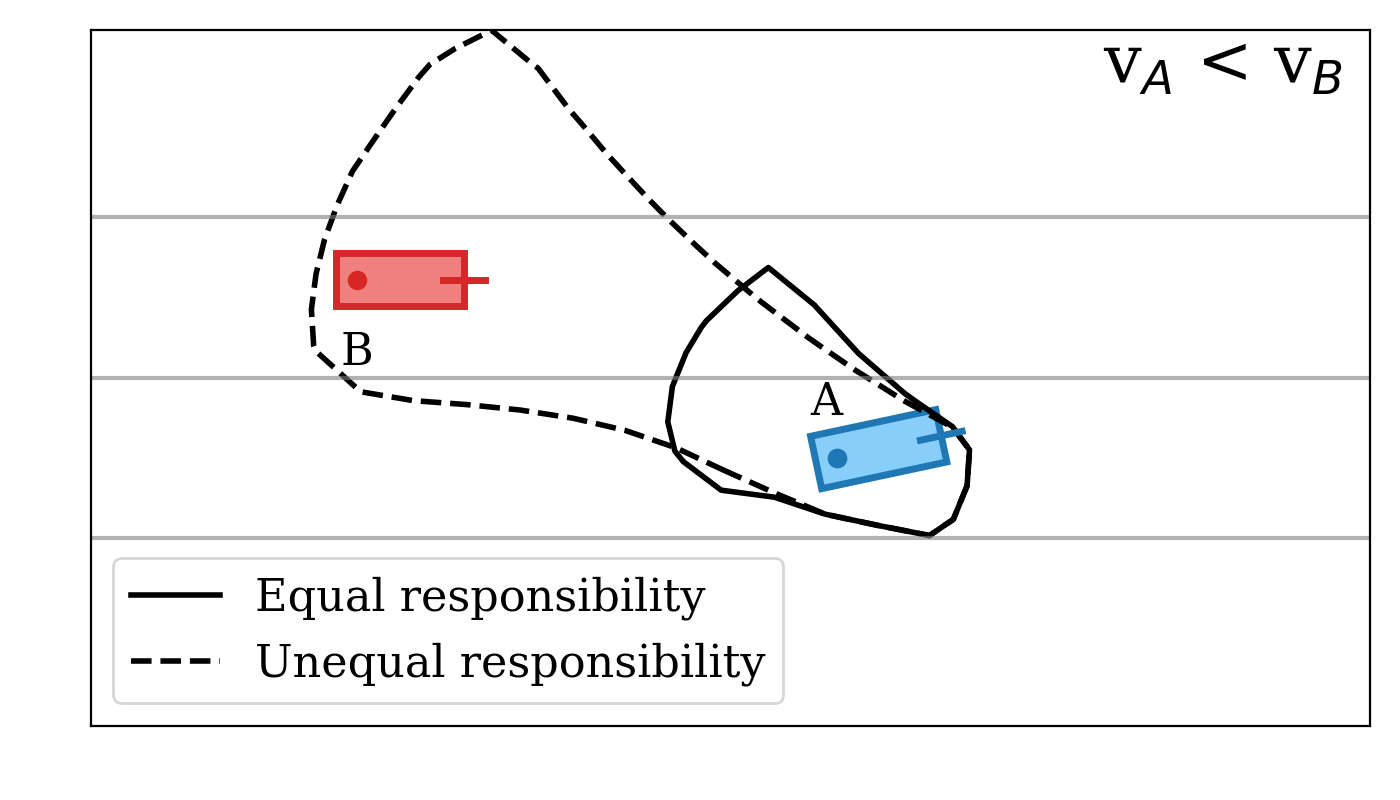}\label{fig:lane merging differing responsibilities}} \hspace{5mm}
     \subfloat[Comparing fixed and state-dependent control sets.]{\includegraphics[width=0.45\textwidth]{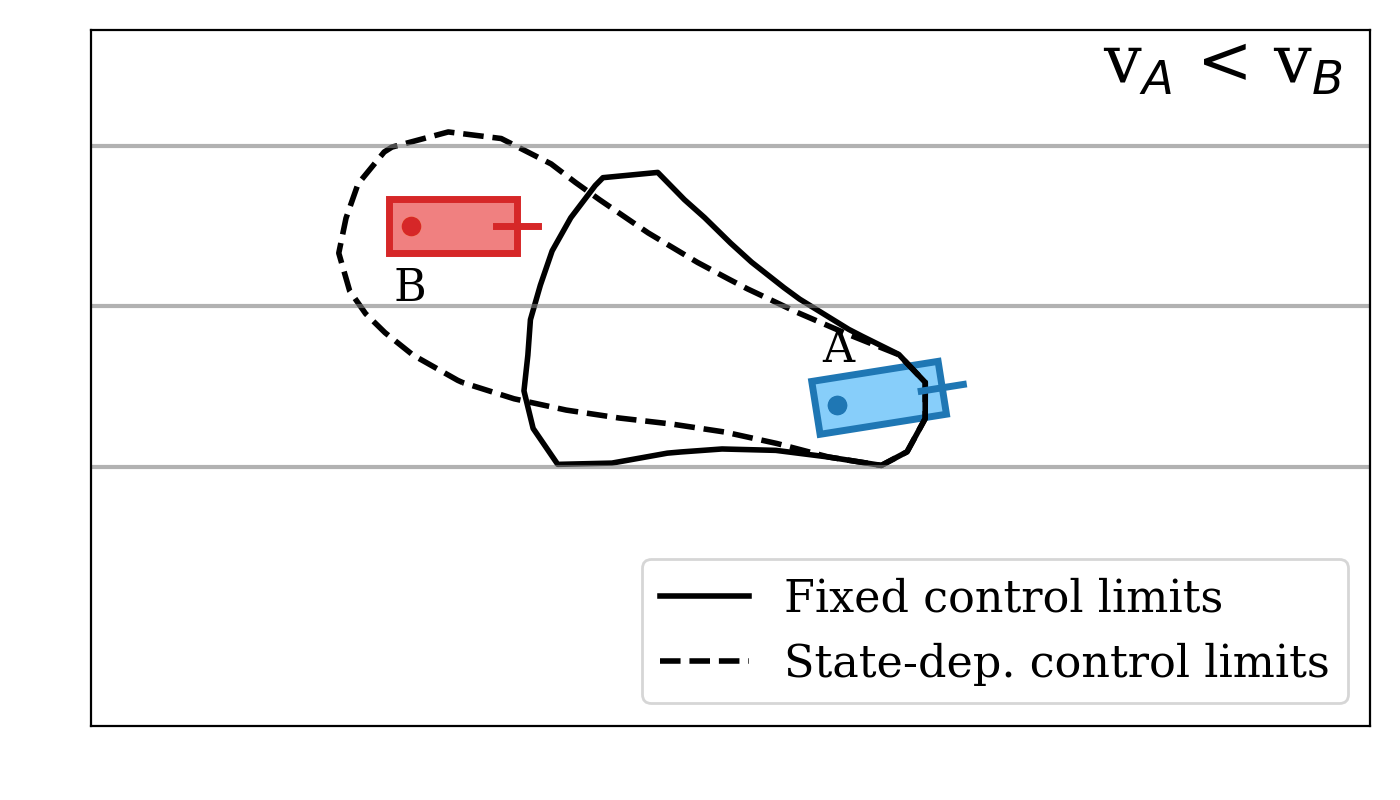}\label{fig:state dependent controls}}
     \caption{Illustrations of the unsafe set $\mathcal{A}(t)$ with various allowable control set restrictions.}
     \label{fig:illustrative examples}
\end{figure}

\noindent{\bf Encoding responsibility via unequal control and velocity limits:}
Consider a simplified highway merging scenario depicted in Figure~\ref{fig:lane merging differing responsibilities}---Agent $A$ is trying to merge into the lane which Agent $B$ is already traveling in. One possibility to encode responsibility is by restricting an agent's control limits. 
Since Agent $B$ is incumbent in the lane and already traveling at high speed, it should not be equally responsible for avoiding a collision, thus its steering and acceleration limits are reduced. Since Agent $A$ is merging and is responsible for yielding, its maximum velocity is limited, but retains full control authority. 
With these choices in control actuation limits, we see that the unsafe sets differ substantially and inherently capture Agent $A$'s responsiblity to ensure sufficient space when cutting in front of a faster moving vehicle.

\noindent{\bf Encoding responsibility via state-dependent control limits:} Alternatively, we can consider control limits as a function of state, specifically, velocity. In Figure~\ref{fig:state dependent controls}, we reduce each agent's steering limits by a factor of $\alpha=\gamma + (1 - \frac{v_i-v_{\min}}{v_{\max}-v_{\min}})(1-\gamma)$, and acceleration limits by $\beta=\gamma +  \frac{v_i-v_{\min}}{v_{\max}-v_{\min}}(1-\gamma)$, where  $\gamma=0.2$ and $v_i$ is the velocity of Agent $i$. With these state-dependent control limits, we observe noticeably different unsafe sets compared to using statically fixed control limits. We handcrafted state-dependent control limits in this example, but we envision that the mapping from state to control bounds can be \textit{learned from data}, which in turn could capture responsibility and context.

\section{Conclusion}

Towards confident deployment and widespread adoption of AVs, stakeholders have aimed to establish a foothold of assurances in the form of fundamental safety concepts. There is still a ways to go, however, in establishing consensus on how to assess safety even in limited/abstracted scenarios, let alone more complicated ones incorporating notions of agent responsibility subject to surrounding context. We have proposed HJ reachability as a unifying mathematical framework for describing a family of safety concepts, including those that already exist, to help stakeholders compare and contrast safety concepts and converge to common ground. Moreover we contend that this framework offers inductive bias amenable to learning data-driven safety concepts that capture more nuanced scenarios, thereby expanding the domain of safety concept applicability.

\vspace{5mm}
{\small \noindent \textbf{Acknowledgements.} The authors would like to thank Boris Ivanovic for his helpful comments on the manuscript draft, and Yunfei Shi, Julia Ng, and David Nister for their insightful discussions.}

\bibliographystyle{unsrt}
\bibliographystyle{apalike}
{
\renewcommand{\baselinestretch}{0.90} 
\bibliography{references_initials}  %
}

\end{document}